\documentclass{article}

\usepackage{xcolor}
\usepackage{graphicx}
\usepackage{url}
\usepackage{amsmath,amssymb}
\usepackage{color,soul}

\usepackage{algorithm}
\usepackage[noend]{algpseudocode}
\usepackage{xr}
\usepackage[final]{corl_2018} 

\title{Integrating kinematics and environment context into deep inverse reinforcement learning for predicting off-road vehicle trajectories}

\author{
  Yanfu Zhang\thanks{Corresponding author: \url{zhangya@yamaha-motor.co.jp}}\\
  Yamaha Motor Co. Ltd.\\
  \And
  Wenshan Wang\\
  Carnegie Mellon University\\
  \And
  Rogerio Bonatti\\
  Carnegie Mellon University\\
  \And
  Daniel Maturana\\
  Carnegie Mellon University\\
  \And
  Sebastian Scherer\\
  Carnegie Mellon University\\
}

\begin{document}
\maketitle


\begin{abstract}

Predicting the motion of a mobile agent from a third-person perspective is an important component for many robotics applications, such as autonomous navigation and tracking. With accurate motion prediction of other agents, robots can plan for more intelligent behaviors to achieve specified objectives, instead of acting in a purely reactive way. Previous work addresses motion prediction by either only filtering kinematics, or using hand-designed and learned representations of the environment. Instead of separating kinematic and environmental context, we propose a novel approach to integrate both into an inverse reinforcement learning (IRL) framework for trajectory prediction. Instead of exponentially increasing the state-space complexity with kinematics, we propose a two-stage neural network architecture that considers motion and environment together to recover the reward function. The first-stage network learns feature representations of the environment using low-level LiDAR statistics and the second-stage network combines those learned features with kinematics data. We collected over 30 km of off-road driving data and validated experimentally that our method can effectively extract useful environmental and kinematic features. We generate accurate predictions of the distribution of future trajectories of the vehicle, encoding complex behaviors such as multi-modal distributions at road intersections, and even show different predictions at the same intersection depending on the vehicle's speed.

\end{abstract}


\section{Introduction}
Most autonomous navigation and tracking applications include interactions with other agents in the world. If a robot can accurately forecast the motion of external agents, it can plan intelligent behaviors, instead of having purely reactive interactions. For example, to avoid collisions with other drivers on a road, an autonomous car should be able to model future trajectories of other vehicles, and a robot manipulator that interacts with humans should anticipate a person's behavior to avoid dangerous situations.

Typically, the motion prediction problem is addressed using filtering-based methods \cite{thrun2005probabilistic}, which use specific kinematic and observation models to estimate the states of the subject, such as position, orientation, and velocity. However, predicting the motion of an agent considering only kinematics is a incomplete model of the true behavior, which also depends on the surrounding environment. Although human experts can manually design functions that model the agent's interactions with objects, this process is usually time-consuming and offers weak generalization to new environments. 

Recently, imitation learning based approaches showed the possibility of learning the complex agent-environment interaction~\cite{kitani2012activity, wulfmeier2017large}. In particular, inverse reinforcement learning (IRL) techniques can recover a complex reward structure using expert demonstrations, and offer higher robustness to generalization than manually-designing functions or supervised learning \cite{osa2018algorithmic}. Although the first IRL reward structures were linear \cite{abbeel2004apprenticeship,ratliff2006maximum,ziebart2008maximum}, recent work \cite{wulfmeier2017large,finn2016guided} improved the reward model complexity.

A challenge in IRL comes from the choice of representation of the agent's state. Adding more dimensions to the state such as velocities and higher-order derivatives generally improves the fidelity of the model, but extra dimensions come with an exponential increase in computation complexity \cite{sutton1998reinforcement}. Finn \cite{finn2016guided} used importance sampling to reduce computation time when considering positions and velocities as the robot state. But in previous IRL work for ground vehicles~\cite{wulfmeier2017large} does not consider kinematics when computing reward maps, using only position as the state.

Our key insight is that one can lower the exponential complexity increase of incorporating kinematics into the state-space if kinematic data is used instead during feature extraction. We offer three major contributions: 1) We propose an improvement on existing deep IRL frameworks, incorporating both kinematic and environmental context to predict trajectories from raw sensory input; 2) We train and verify the proposed method on a custom off-road driving dataset; 3) We qualitatively and quantitatively compare our prediction results with baseline methods such as extended Kalman filters (EKF) and direct behavior cloning (BC).

\begin{figure}[t]
    \centering
    \includegraphics[width=1.0\textwidth]{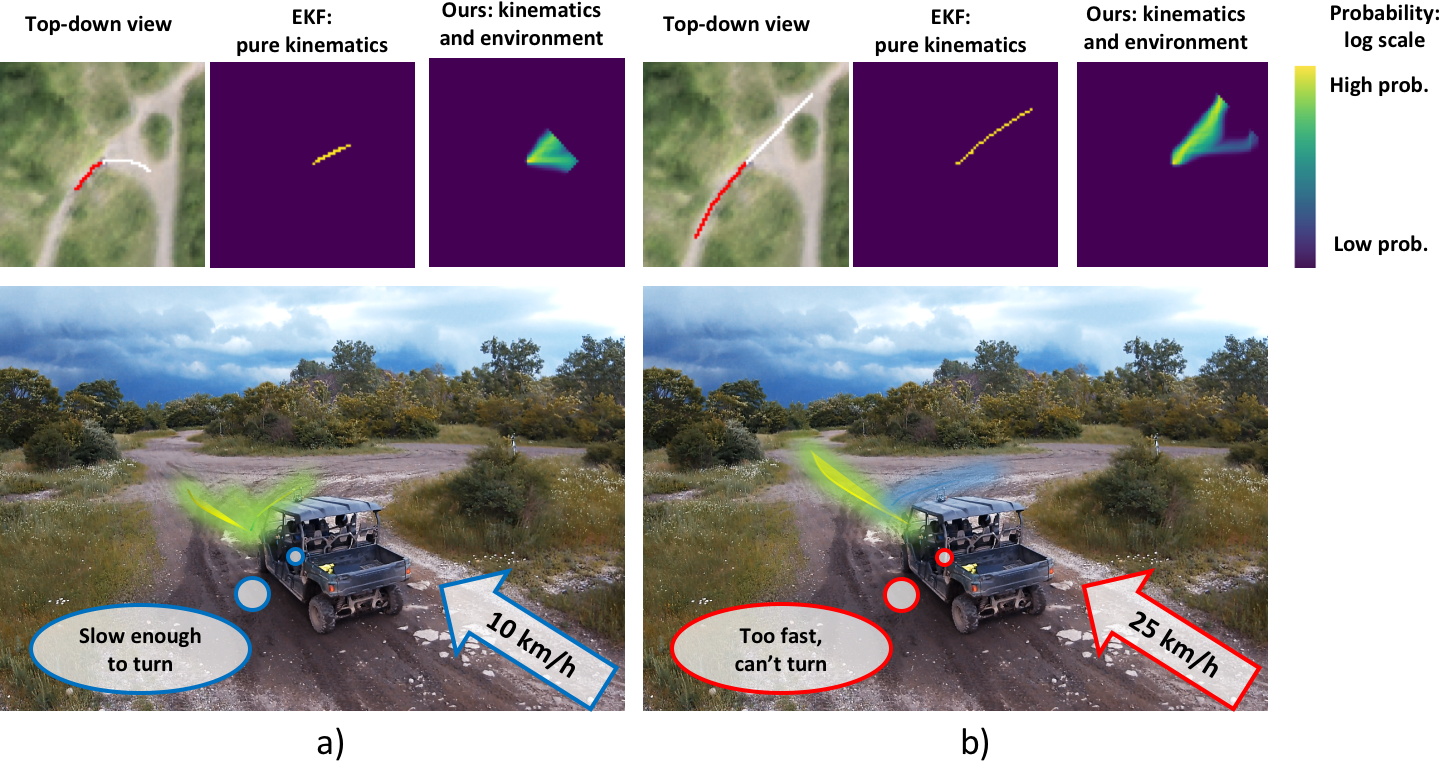}
    \caption{Trajectory prediction results on the same environment comparing different speeds. Slow speeds show stronger multi-modal prediction (a), while a fast vehicle is more likely to just continue going straight (b). The top-down view shows the trajectory of the past five seconds (red) and ground truth (white). We compare our method with an extended Kalman filter prediction.}
    \label{fig:key_result}
\end{figure}


\section{Related work}
\label{sec:relatedwork}

The problem of predicting an agent's motion can be approached using different frameworks depending on the system's objective. A Kalman filter estimates linear and angular velocities, then forward-propagates the motion~\cite{thrun2005probabilistic}. The filter ignores environmental context, and provides a uni-modal distribution over the space of possible trajectories. To model the agent's interactions with objects in the world, different methods manually design cost functions~\cite{choset2005principles, urmson2009autonomous, thrun2006stanley}. However, to overcome the issue of human fine-tuning and low generalization, one can explore the idea of imitation learning.

BC methods to model driving date as far back as 1989, when Pomerleau~\cite{pomerleau1989alvinn} used supervised learning to map the current first-person image from a car to the desired steering angle, and methods improved significantly over time~\cite{bojarski2016end}. A drawback of BC is that in practice it is based on supervised learning with \textit{non-i.i.d} samples due to the sequential aspect of data, therefore matching the training and test set distributions can be a challenge, hindering generalizability~\cite{osa2018algorithmic}, despite techniques like DAgger~\cite{ross2011reduction} are developed to mitigate this problem.

IRL can be used to improve generalization of a policy. IRL's objective is to recover the reward function that the agent optimizes \cite{osa2018algorithmic}. While early work on IRL modeled rewards as linearly dependent on features \cite{abbeel2004apprenticeship, ratliff2006maximum, ziebart2008maximum, kitani2012activity}, more complex models followed \cite{finn2016guided, ho2016generative, levine2011nonlinear,wulfmeier2017large}. 

Closest to our work, Wulfmeier et al. \cite{wulfmeier2017large} use a neural network to model the mapping from sensor measurements coming from expert driving demonstrations to a reward map. A motion planner can then use the reward map to generate a path between desired start and goal locations. However, this approach ignores the role of dynamics in the forecast, considering only positions in the state-space. Adding extra dimensions to the state exponentially increases computational complexity for the task \cite{sutton1998reinforcement}. Thus, the training process slows down significantly and requires larger amounts of demonstrations for converge of a policy. Finn \cite{finn2016guided} attempts to overcome complexity by using importance sampling to compute state visitation frequencies from the learned policy. In this work, however, we argue that kinematics can instead be successfully used in the feature extraction process. Therefore, we propose a non-linear reward model that depends on both kinematics and environmental features, while keeping a small state-space dimension.


\section{Approach}
\label{sec:approach}
We frame the problem of vehicle trajectory prediction with max-entropy IRL~\citep{ziebart2008maximum}. In order to incorporate both environment and kinematic context, a two-stage convolutional neural network (CNN) architecture is utilized to approximate the underlying reward function.

\subsection{Problem formulation}

Our objective is to learn to predict the target's future path $\zeta$ or probability distribution of future path $p(\zeta_i)$. We use a set of demonstration trajectories $D=\{\zeta_0, \zeta_1, ... \zeta_m\}$, which are collected in various environments. The trajectory $\zeta$ is defined as a sequence of states $(s_0, s_1,..., s_n)$, where $s=(x,y)$ represents the target's 2D location in world frame. Each state $s$ contains features $\phi(s)$, obtained from sensory inputs. To predict the trajectory, one can directly estimate the state sequence, or predict the action sequence $(a_0, a_1, a_2, a_3)$, where action $a$ is a discretized motion (up, down, left, right). 

In BC, we can find a mapping from state and features to action: $\pi:s, \phi(s) \mapsto{a}$ \cite{osa2018algorithmic}. In IRL, however, the goal is to recover the agent's hidden reward function $R(\phi(s))$, so as to maximize the probability of the demonstrated trajectories. Once the hidden reward is inferred, the probabilities of future trajectories can be computed. 

\subsection{Maximum entropy deep IRL}
Here we follow the maximum entropy IRL formulation~\citep{ziebart2008maximum}, treating trajectories with higher rewards as exponentially more likely. As shown by \cite{wulfmeier2017large,finn2016guided}, we approximate the reward function using a deep neural network $R(\phi(s))=f(\phi(s);\theta)$. We define the reward of a trajectory $\zeta$ as the accumulative reward over all states in that trajectory $R(\zeta)=\sum_{s_i\in\zeta}{f(\phi(s_i);\theta)}$ Under the maximum entropy assumption, we derive the probability of trajectory $\zeta_{i}$ given the reward function as  $P(\zeta_i|\theta)=\frac{1}{Z(\theta)}{\exp{R(\zeta_i)}}$,
where $Z(\theta)=\sum_{\zeta_j \in D}{\exp{R(\zeta_j)}}$ is the partition function over all possible trajectories. Then we try to maximize the following log likelihood of the demonstrated trajectories \cite{wulfmeier2017large}:

$$\mathcal{L}(\theta)=\log \prod_{\zeta_i \in D} {P}(\zeta_i;\theta) = \sum_{\zeta_i\in{D}}{R(\zeta_i)}-\log{M\sum_{\zeta_j}{\exp(R(\zeta_j;\theta))}}
\Rightarrow
\frac{\partial{\mathcal{L(\theta)}}}{\partial{\theta}}=(\mu_D-\mathbb{E}[\mu])\frac{\partial{f}}{\partial{\theta}}
$$

Where $\mu_D$ and $\mathbb{E}[\mu]$ are the state visitation frequencies (SVF) from the demonstrated trajectories and from the inferred reward function respectively. After the reward network update, we use standard value iteration to solve the forward RL problem in the loop, as shown in Alg.~1. During value iteration, to speed up the convergence rate we artificially increase the probability of the most likely action being chosen, in what we call \textit{annealed softmax} in Alg.~2.

\begin{algorithm}
\caption{Deep maximum entropy IRL with two-stage network architecture}\label{euclid}
\textbf{Input:} D, S, A, T, $\gamma$, $\alpha$\\
\textbf{Output:} network parameters $\theta$
\begin{algorithmic}[1]
\label{alg:main}
\State Initialize network parameters $\theta$ randomly
\For{iteration $i=1$ to $N$}
  \State sample demonstration batch $D_{batch} \subset D$
  \State $R_i(\phi(s)) \gets f(\phi(s);\theta_i)$ for $\forall s \in S$   \Comment{Forward reward network}
  \State $\pi_i \gets value\_iteration (R_i,S,A,T,\gamma)$ \Comment{Planning step}
  \State $\mathbb{E}[\mu_i] \gets compute\_svf(\pi_i,S,A,T)$
  \State $\frac{\partial{\mathcal{L(\theta_i)}}}{\partial{R}} \gets \mu_D-\mathbb{E}[\mu_i] $ \Comment{Gradient calculation}
  \State $\theta_{i+1} \gets back\_propagate(f,\theta_i, \frac{\partial{\mathcal{L(\theta_i)}}}{\partial{R}}, \alpha)$ \Comment{Parameter update}
\EndFor
\State \Return $\theta$
\end{algorithmic}
\end{algorithm}

\begin{algorithm}
\caption{Value iteration}\label{euclid}
\textbf{Input:} R, S, A, T, $\gamma$\\
\textbf{Output:} $\pi$
\begin{algorithmic}[1]
\label{alg:value_iter}
\State Initialize values $V(s)=-\inf$
\Repeat{}
  \State $V_t(s) = V(s)$ \Comment{No hard reset for $V(s_{goal})$}
  \State $Q(s,a) = r(s,a) + E_{T(s,a,s')}[V(s')]$
  \State $V(s) = max(Q_i(s,a))$
\Until{ $max_s(V(s)-V_t(s))<\epsilon$}
\State \Return $\pi(a|s) = annealed\_softmax(Q(s,a))$
\end{algorithmic}
\end{algorithm}

\subsection{Incorporating kinematics into the reward: two-stage network architecture}

\begin{figure}[b]
    \centering
    \includegraphics[width=0.8\textwidth]{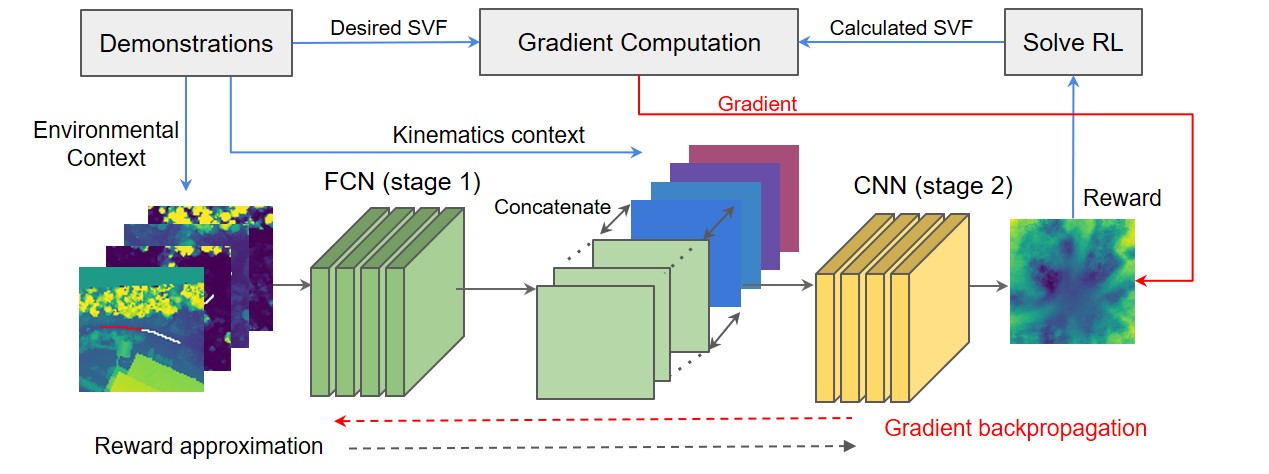}
    \caption{Proposed two-stage network architecture and training procedures. Environmental context is the input to the first-stage network, and the resulting feature maps are concatenated with the kinematics context. The reward approximation is the output of the second-stage network, and the SVF from the learned policy is used to compute gradients for backpropagation.}
    \label{fig:two_stage}
\end{figure}

Intuitively, when humans predict vehicle motion from a third-person perspective we primarily reason about two aspects: the surrounding environment and the vehicle's motion so far. We implicitly evaluate traversability of the terrain, the location of obstacles, and extrapolate the past vehicle motion to infer where turning will be necessary and/or possible. If given enough observations, we can even infer more subtle cues such as driver preferences and a specific motion model for the vehicle.

Inspired by this intuition, we designed a two-stage network architecture to reason about environmental and kinematic context when computing rewards (Figure~\ref{fig:two_stage}). In the first stage, we adopt a four-layer fully convolutional network (FCN)~\citep{long2015fully} structure which takes colored point cloud statistics as inputs, and outputs features extracted purely from the environmental context. Instead of using encoder-decoder approaches~\citep{long2015fully,badrinarayanan2017segnet,paszke2016enet}, which inevitably lose spatial information due to the down-sampling stages, we adopt dilated convolutional layers, which systematically aggregate multi-scale context without losing spatial information~\citep{yu2015multi}. The approximate receptive field of our dilated network is $20 \times 20$ pixels, which we estimated to be sufficient to extract the relevant environmental context. The final output of the dilated network is 25 feature maps, a number experimentally observed to contain sufficient information representing spatial context without excessive redundancy and sparsity.

As inputs to the second stage, we concatenate the output from the first stage with two feature maps encoding positional information and three feature maps representing kinematic information from the past trajectory, as follows.
The first two feature maps encode, for each grid cell, the $x$ and $y$ position of the grid cell in a vehicle-centered, world-aligned frame. These feature maps are independent of the trajectories, but convey absolute position information to FCN, which is translation-invariant\footnote{An alternative way of encoding absolute spatial position would be using a fully-connected network, but this will result in a substantial increase in the number of parameters.}. Then, for each training sample, past trajectory information is encoded with three feature maps $\phi_e(\zeta)=[\Delta x, \Delta y, \kappa]$. $\Delta x$ and $\Delta y$ represent the vehicle's past velocity, discretized to the four cardinal directions, and with a normalized magnitude proportional to its absolute speed. $\kappa$ encodes the trajectory curvature, which is related to angular velocity. $[\Delta x, \Delta y, \kappa]$ are estimated from the past five seconds of vehicle motion and respectively constant across the whole feature map. Finally, they are empirically normalized to a small finite range to aid training stability. More implementation details can be found in Appendix.


\section{Experiments}
\label{sec:experiments}
In this section we present our approach for dataset collection, baseline design, definition of metrics, prediction results, and discussion on learned results.

\subsection{Off-road driving dataset collection}
\begin{figure}[h]
    \centering
    \includegraphics[width=0.5\textwidth]{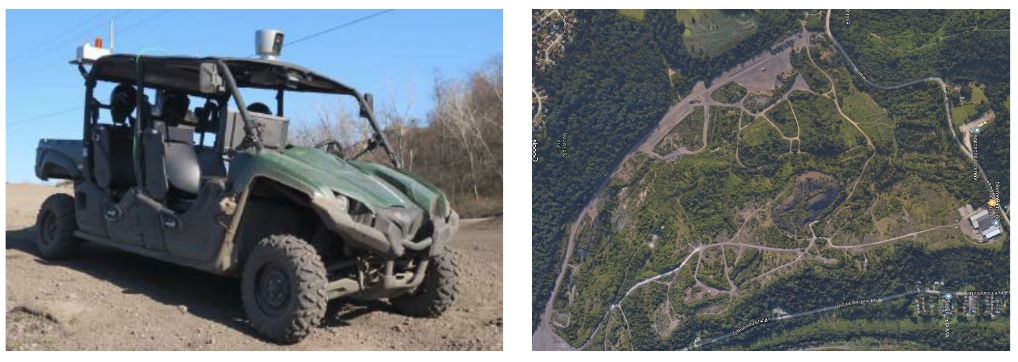}
    \caption{Vehicle platform, along with satellite view of the off-road course used for about 30 km of data collection.}
    \label{fig:viking}
\end{figure}

The off-road driving dataset was collected in a test site with roughly 400 acres involving more than 5 different drivers. The vehicle platform, a modified All Terrain Vehicle (ATV), is shown in Figure~\ref{fig:viking}. A total of over 1000 trajectories are included for the dataset, with average length about 20-40 m long, in about 30 km of driving demonstrations.

For each demonstration we create a high-precision, colored local point cloud map. We then convert the point cloud into a 2D grid with statistics for each cell: maximum height, height variance, and mean RGB values, and extract the vehicle past trajectory and ground truth prediction using high precision RTK-GNSS/INS which can provide position data with centimeter level accuracy.

\begin{figure}[t]
    \centering
    \includegraphics[width=0.8\textwidth]{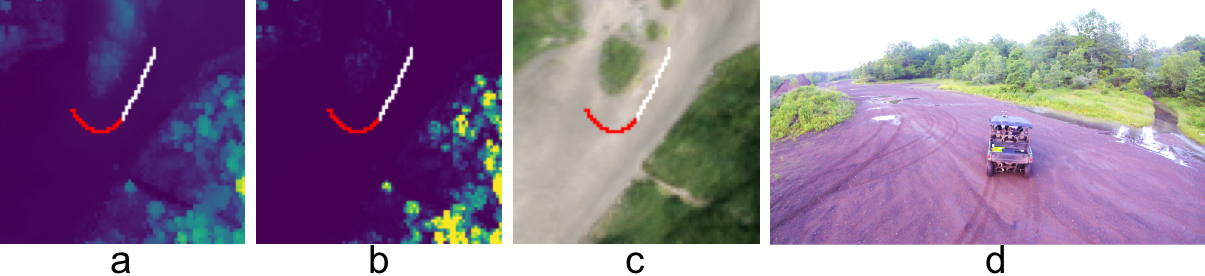}
    \caption{Visualization of all channels of the input to the reward network. a) Maximum height, b) Height variance, c) RGB, d) Third-person view of the environment (not an input). Red indicates the trajectory history, which is used to compute kinematic features, and ground truth future path is shown in white.}
    \label{fig:sample_data}
\end{figure}

The reason behind our choice of features is that in off-road environments, a combination of geometry and color statistics can provide useful information regarding terrain traversability. For example, one can intuitively categorize regions with high variance in height to be rough and non-traversable. However, combining geometry with color can be even more informative: an area that is simultaneously rough but light green is probably just a region of tall grasses, over which the vehicle can easily drive. An example of data from our dataset of demonstrations is shown in Figure~\ref{fig:sample_data}. 

\subsection{Metrics and baselines}
We employ two metrics for quantitative evaluation: the Negative Log-Likelihood (NLL) of the demonstration data and the Hausdorff Distance (HD) \cite{kitani2012activity}. The NLL metric computes the log-likelihood of the demonstration trajectory $\zeta$ under the learned policy $\pi(a|s)$. We normalize NLL by the demonstration trajectory length. The HD metric represents a spatial similarity between expert demonstrations and trajectories sampled with the learned policy. To compute it, we use the average HD between the demonstration trajectory and 1000 trajectories randomly sampled from the learned policy. 

We selected three baseline methods for comparison: 1) EKF with a kinematic bicycle model; 2) BC technique that uses supervised learning to learn a policy mapping both environment and kinematic inputs to an action; and 3) deep IRL method considering only environment input to compute features, with no kinematics.

\subsection{Trajectory prediction results}

\textit{Qualitative evaluation}

\begin{figure}[b]
    \centering
    \includegraphics[width=0.8\textwidth]{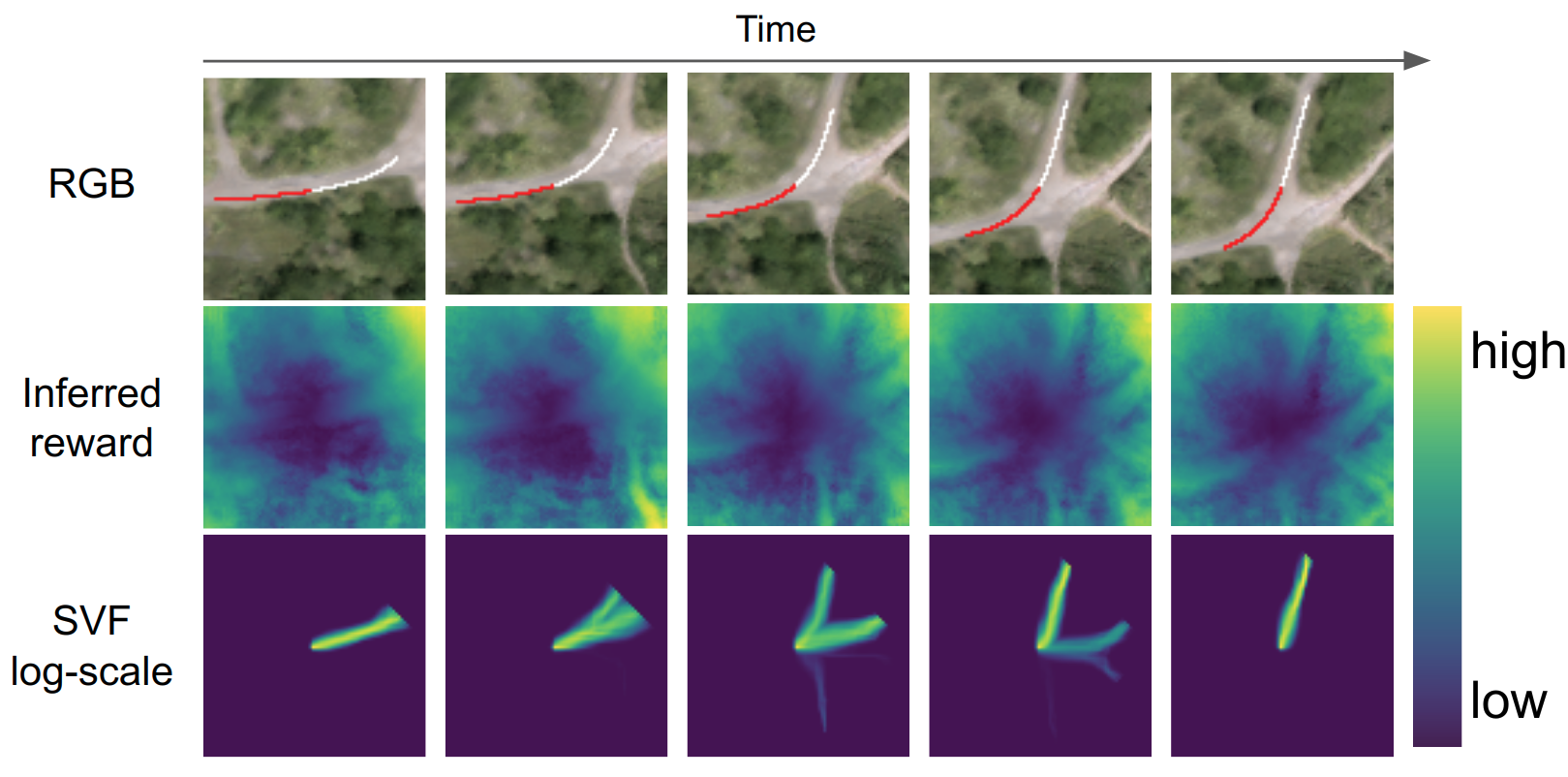}
    \caption{Time-lapse of inferred reward map and trajectory forecast distributions in an off-road trail test set. Forecasts remain on trails, and show multi-modal distributions at intersections.}
    \label{fig:qualitative_result}
\end{figure}

\begin{figure}[t]
    \centering
    \includegraphics[width=0.8\textwidth]{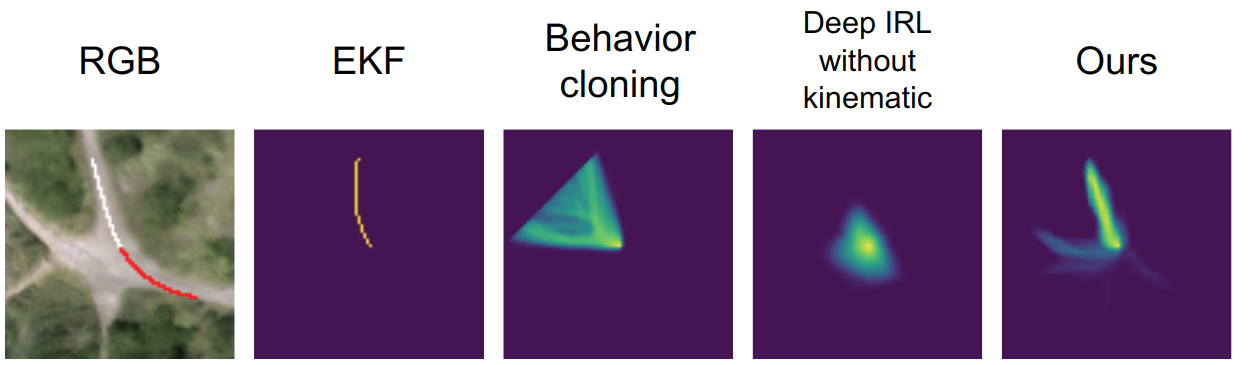}
    \caption{Comparison of trajectory predictions for different baselines in test set. The EKF ignores any environmental context, and behavior cloning fails to learn a policy that avoids obstacles. Without kinematic features, deep IRL has no notion of where the vehicle is going, and just forecasts a diffuse motion on the large open section of trail around the actor's position. Our method is able to capture both kinematics and environmental cues, inferring multiple plausible paths.}
    \label{fig:compare}
\end{figure}

We show key experimental results in Figure~\ref{fig:qualitative_result}, and a graphical comparison with baselines in Figure~\ref{fig:compare}. We verify that the forecasts based on the inferred reward map, match the expected behaviors of a human driver, with trajectory probabilities concentrated on trails or traversable paths. In addition, multi-modal behaviors are present at intersections and when approaching open areas.

In Figure~\ref{fig:key_result} we see how the trajectory prediction behaves on the same environments, for different agent kinematics. Our predictions are surprisingly intuitive: at low speed the forecast shows higher likelihood of taking turns at an intersection, while at high speed it predicts that the vehicle will probably keep driving straight along the trail. More visualizations can be found in the demo video\footnote{\url{https://github.com/yfzhang/vehicle-motion-forecasting}}.

\textit{Quantitative evaluation}

We compare our method with four baselines in the off-road driving dataset, shown in Table~\ref{table:acc_compare}. Our method has the best prediction results on the test set using both metrics. We observe that the EKF obtains surprisingly good results for HD; we believe that this is due to a relatively large proportion of straight or near-straight trajectories in the dataset, which EKF can predict well. We also observe a comparatively poor performance of Deep IRL without kinematic features. Examination of the predictions suggests that without past kinematic context, in many open scenarios this method can only predict diffuse, highly uncertain future motions, as shown in Fig.~\ref{fig:compare}. On the other hand, our method, which can exploit both environmental and kinematic context, obtains the best results.

\begin{table}[h]
\begin{center}
\begin{tabular}{c c c c c c}
\hline
Method & EKF & Behavior cloning & Random & Deep IRL without kinematics  & Ours \\
\hline
NLL & N.A. & 0.87 & 1.35 & 1.33 & \textbf{0.69} \\
HD & 9.12 & 10.54 & 25.62 & 25.46  & \textbf{6.71} \\
\hline
\end{tabular}
\end{center}
\caption{Prediction performance comparison on the test set using NLL and HD. For both NLL and HD, lower numbers represent better predictions. Our method obtains the best results in both evaluation metrics.}
\label{table:acc_compare}
\end{table}

\subsection{What is learned?}
\begin{figure}[h]
    \centering
    \includegraphics[width=1.0\textwidth]{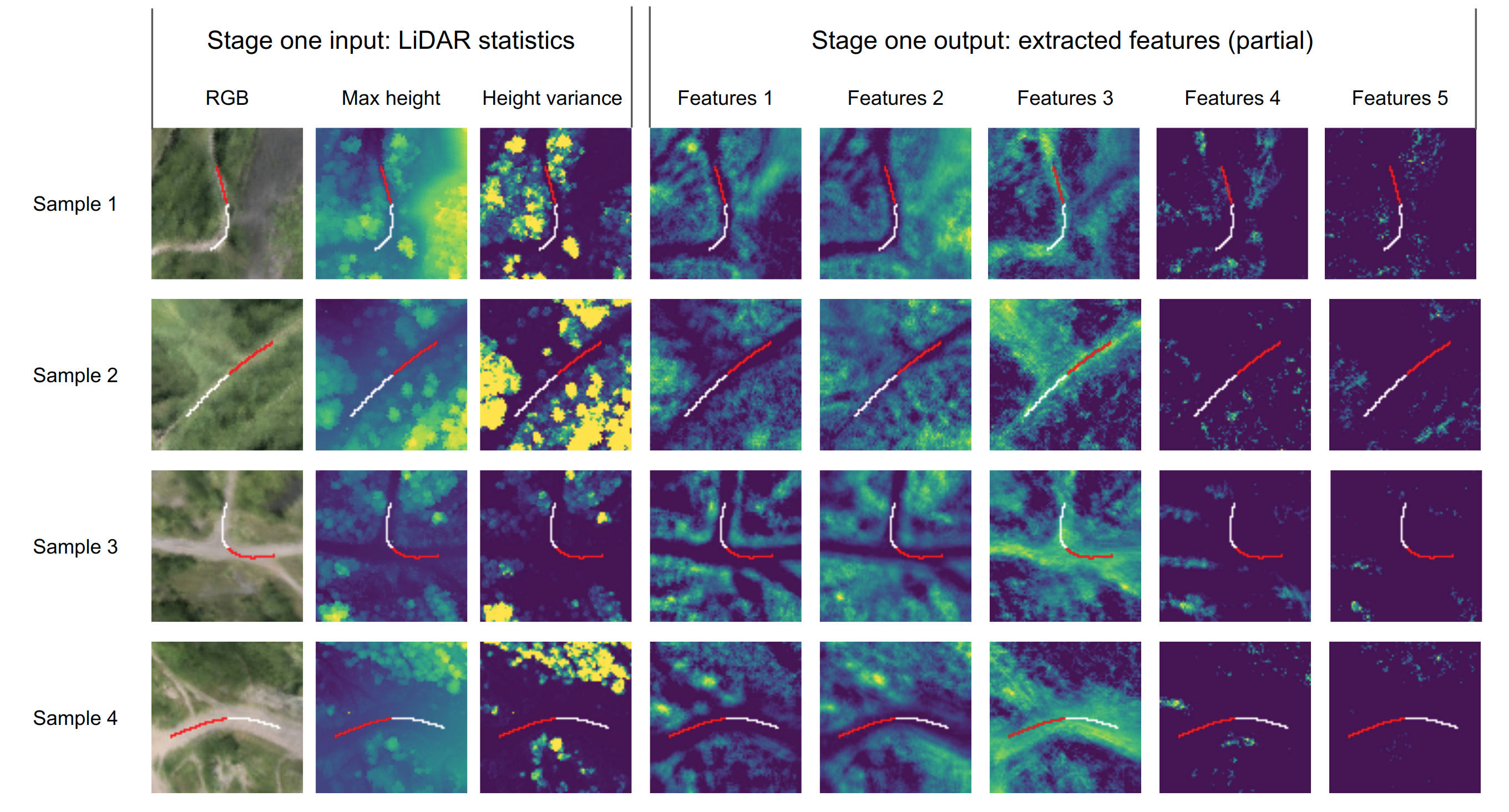}
    \caption{Visualization of select feature maps from the first-stage network output. Features 1-3 have dense output patterns that visibly correlate to the trail region. Features 4-5 show a relatively sparse output, possibly encoding specific LiDAR statistics.}
    \label{fig:feat_out}
\end{figure}

\begin{figure}[h]
    \centering
    \includegraphics[width=0.8\textwidth]{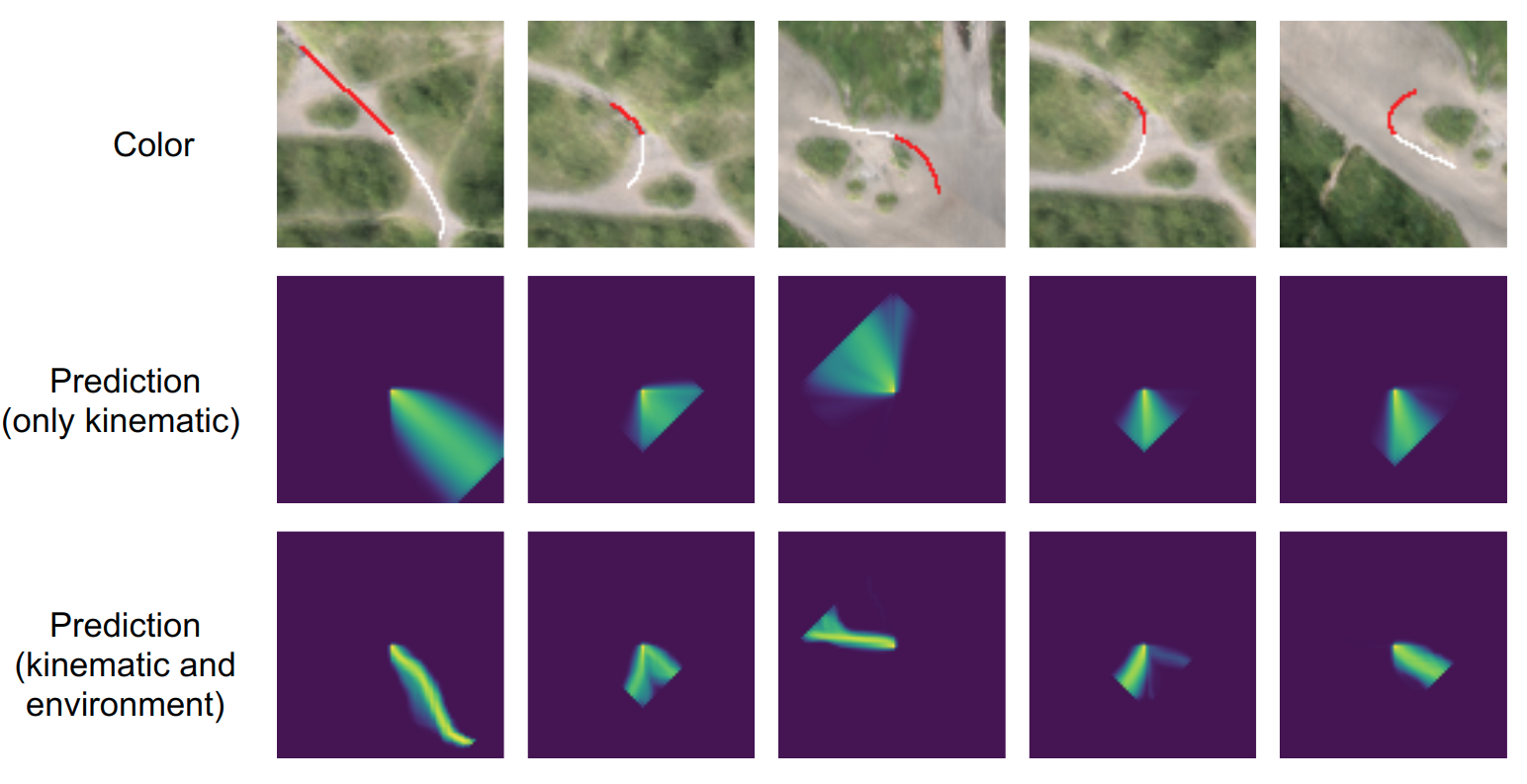}
    \caption{Visualization of trajectory probabilities with and without environment context. The first row shows past trajectories (red) and ground-truth (white). The second row shows predictions using our method where LiDAR features were replaced by a constant value in all cells, and we can compare with predictions that incorporate both kinematics and environment in the third row. The network learns a forward motion model of the car with an uncertainty cone roughly oriented towards the direction of motion, and that varies with respect to vehicle speed.}
    \label{fig:kine_out}
\end{figure}

We try to build an understanding of what is being learned by plotting representative outputs from different stages of the network. Shown in Figure~\ref{fig:feat_out}, the first-stage is network apparently encoding the traversable trail based on colored point cloud input. Shown in Figure~\ref{fig:kine_out}, the second stage of the network learns an effective forward motion model of the car with an uncertainty cone roughly oriented towards the direction of motion.

\section{Discussion and future work}
\label{sec:conclusion}
In this work we proposed and validated a deep IRL approach that integrates kinematic and  environmental context to learn a reward structure for driving predictions in off-road environments. We show that our approach out-performs baselines such as EKF, BC, and a deep IRL approach without considering kinematics. 

Our two-stage network architecture used for the reward approximation can mitigate the computational complexity of adding kinematics as extra dimensions to the state-space; we instead add kinematics in the feature extraction process. Based on our findings, we see future work in a few directions:

\textit{Dynamic environment}: Our current approach can not handle dynamic environment which contains other moving agents, such as pedestrians and other vehicles. As an extension to the current framework, we can take a history of visual and LiDAR data as the environmental input and this temporal information can be used to infer the motion of additional moving agents in the environment. A recurrent network probably can fuse this sequential information and approximate the reward structure.

\textit{Improve kinematics model}: Our kinematic features come from a fixed-time window from the past trajectory. When the driver does a sudden motion, using a fixed-time window is not ideal, because under sudden movements only a smaller time segment should be considered to forecast where the vehicle will go. We envision improvements in the second-stage network, where the network can learn how to select adaptive time windows for prediction, probably using recurrent networks.
 


\clearpage
\acknowledgments{This research is funded by Yamaha Motor Co. Ltd. We thank Lentin Joseph, Masashi Uenoyama, Yuji Hiramatsu for data collection and the anonymous reviewers for providing feedback.}


\bibliography{example}  

\clearpage
\appendix
\section{Additional experimental detail}
In this section, we provide additional details for all experiments, including data pre-processing, baseline implementation, network parameters, and training hyperparameters.
\subsection{Data pre-processing}
\textit{Environment input}. We fuse camera data with LiDAR data to generate colored point cloud. For each demonstration, we create a local 3D point cloud map via registration and further compress it into a 2D grid ($80$ m $\times$ $80$ m, with $1$ m resolution) with five channels: max-height, height variance, mean red, mean green, and mean blue.

\begin{figure}[h]
    \centering
    \includegraphics[width=0.8\textwidth]{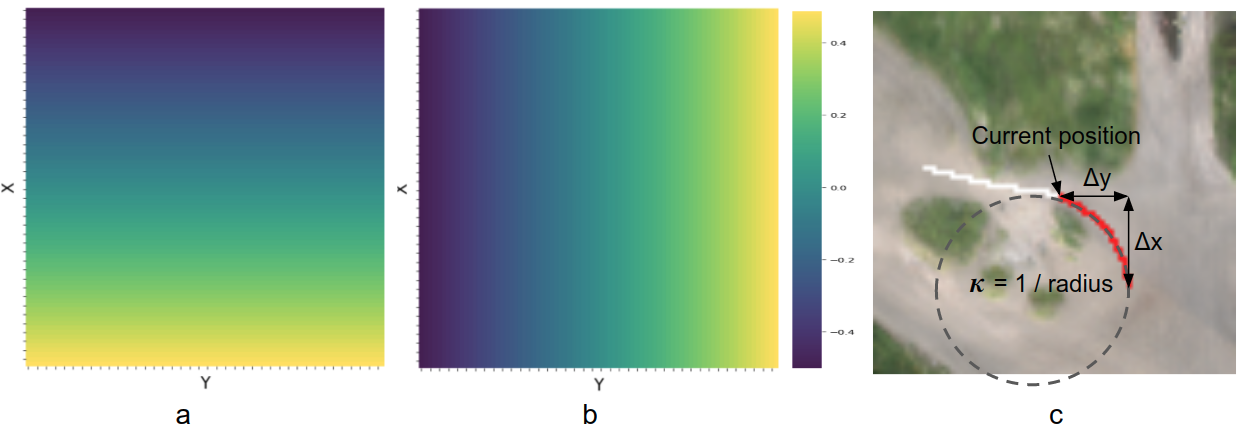}
    \caption{Illustration of the kinematic input. a) and b) are two layers encoding the [x, y] positional information. They are concatenated with other feature maps to convey positional information to convolution and break translation-invariance. c) illustrates how kinematic data $[\Delta x, \Delta y, \kappa]$ is calculated from the past trajectory which is shown in red.}
    \label{fig:kine_illustrate}
\end{figure}

\textit{Kinematic input}. We have 5 channels encoding positional and kinematic information. The first two channels encode, for each grid, the positional information in a vehicle-centered, world-aligned frame as shown in Figure~\ref{fig:kine_illustrate}. They provide positional information to convolutional filters and break translation-invariance. The other three channels encode the kinematic information $[\Delta x, \Delta y, \kappa]$, which are estimated from the past trajectory as illustrated in Figure~\ref{fig:kine_illustrate} and are uniform over the whole feature map. $[\Delta x, \Delta y]$ are calculated on the grid-map and therefore are discretized. They approximate the vehicle's past linear velocity. $\kappa$ is estimated by fitting a least square circle to the past trajectory and approximates the trajectory curvature, which is related to angular velocity. All the kinematic input is empirically normalized to aid training stability.

\subsection{Training and implementation details}
We implemented the training and inference pipeline using PyTorch~\footnote{\url{https://pytorch.org/}} and our code and dataset will be made public available\footnote{For code and dataset, see \url{https://github.com/yfzhang/vehicle-motion-forecasting}}. We use a 90/10 split to divide our data into training and test sets. We used the following techniques for training the network:

\textit{Data augmentation}. Augmenting the demonstration data with different rotations is critical because the proposed two-stage network is sensitive to directional bias. We wanted to avoid over-fitting to a particular direction without properly reasoning about kinematics. 

\textit{Demonstration distribution}. The network tends to overfit to straight line predictions if the majority of the demonstrations are straight as well, sometimes ignoring environmental features. Therefore we balanced the distribution of types of trajectories, such as straight lines, small and large curves.

\textit{Simulated annealing}. To improve convergence during training time, we artificially increased the probability of the most likely action being chosen.

\textit{Training in parallel}. Since for every training iteration, RL needs to be performed to compute the gradients, we implemented parallel computation of gradients for each batch of demonstrations. We used batches size of 16 demonstrations. Batch training proved in practice to be more stable than using gradients based on single demonstration.

\subsection{Baseline details}
For all learning-based baselines, we set all training hyperparameters the same as those used in our method, and stop training when it starts to show overfitting in the validation set.

\textit{Deep IRL without kinematics}. We removed the second-stage and kept the first-stage the same as the original network design in our approach. Therefore, there is no kinematic data used in training and testing. The results is as expected that it predicts a distribution concentrated in the path, but diffused over all directions compared to our approach. Because of this omnidirectional diffusion, the quantitative evaluation score of this baseline is very low. The reason why deep IRL without kinematics work for the prior work~\cite{wulfmeier2017large} is that a explicit goal state can be provided in the motion planning problem setting. In both training and testing phases, the value of the goal state can be hardly reset and pulls the optimal policy towards that direction. However, in our problem setting, the aim is to predict the future motion in a finite-time horizon, and there is no explicit goal state.

\textit{BC}. We kept the network architecture the same as that in our approach. But instead of outputting the reward values, we set the last layer to output four channels and add a softmax layer. Therefore, the final output becomes the policy and contains the probabilities of four different actions. The network input is also the same as our approach and contains both environment (LiDAR + RGB) and kinematic data.

\textit{EKF}. We adopted an EKF with kinematic bicycle model for state estimation. We use the vehicle position $[x, y]$ observed in the past trajectory as the measurement input and implicitly estimates the vehicle's linear and angular velocity. During the predication phase, we assume the vehicle's linear velocity and front steering angle is unchanged. In the experimental results, we observed that EKF-based prediction performs well when the demonstration trajectory is near-straight or has a constant curvature.

\end{document}